\documentclass[sn-mathphys]{sn-jnl}
\usepackage{fix-cm}
\usepackage{comment}
\usepackage{enumitem}

\begin{document}

\title[Personalized Knowledge Transfer]{Personalized Knowledge Transfer Through Generative AI: Contextualizing Learning to Individual Career Goals}


\author{\fnm{Ronja} \sur{Mehlan}}\email{ronja.mehlan@iu.org}
\author{\fnm{Claudia} \sur{Hess}}\email{claudia.hess@iu.org}
\author{\fnm{Quintus} \sur{Stierstorfer}}\email{quintus.stierstorfer@iu.org}
\author{\fnm{Kristina} \sur{Schaaff}}\email{kristina.schaaff@iu.org}

\affil{\orgdiv{IU International University of Applied Sciences}, \orgaddress{\city{Erfurt}, \country{Germany}}}


\abstract{As artificial intelligence becomes increasingly integrated into digital learning environments, the personalization of learning content to reflect learners’ individual career goals offers promising potential to enhance engagement and long-term motivation. In our study, we investigate how career goal-based content adaptation in learning systems based on generative AI (GenAI) influences learner engagement, satisfaction, and study efficiency. The mixed-methods experiment involved more than 4,000 learners, with one group receiving learning scenarios tailored to their career goals and a control group. Quantitative results show increased session duration, higher satisfaction ratings, and a modest reduction in study duration compared to standard content. Qualitative analysis highlights that learners found the personalized material motivating and practical, enabling deep cognitive engagement and strong identification with the content. These findings underscore the value of aligning educational content with learners’ career goals and suggest that scalable AI personalization can bridge academic knowledge and workplace applicability.}




\maketitle
\section{Introduction}

The global labor market is changing rapidly, driven by digitalization, automation, and evolving demands for new skills. Traditional education systems often struggle to keep pace, relying on rigid curricula and standardized content that fail to meet the diverse and dynamic needs of today’s learners \cite{oecd2019, unesco2023}. 
Personalization in education has long been recognized as a critical factor in improving learning outcomes \citep{vanlehn2011relative}. The traditional model of one-size-fits-all instruction often neglects the diverse needs, preferences, and backgrounds of learners.
In many cases, learners do not see the relevance of academic material for their personal development or future careers, which negatively affects engagement and long-term learning outcomes \cite{oecd2023}.
With the rise of digital education, various adaptive learning systems have attempted to close this gap---yet many remain limited in scope and flexibility.

To address this gap, in our study, we focus on career goal-based personalization---an approach that adapts learning content to align with the professional aspirations of individual learners. For example, the same core concept---such as data interpretation---may be taught through different scenario-based tasks: analyzing patient records for aspiring healthcare professionals, evaluating marketing KPIs for future brand managers, or reviewing logistics reports for those targeting supply chain roles. Research shows that when learners perceive clear connections between their studies and future roles, they are more motivated, persistent, and capable of transferring knowledge into real-world applications \citep{deci2000, eccles2002, dweck2006}.

We use generative AI (GenAI) to implement scalable, context-aware personalization. By aligning scenarios and tasks with learners’ intended career goals, we aim to increase not only motivation but also depth of processing, retention, and perceived relevance. Recent advances in GenAI make it possible to generate industry-specific examples and role-relevant challenges in real-time, offering a more meaningful and cognitively stimulating learning experience~\citep{kasneci2023, ng2024}.

In our study, we explore how aligning educational content with learners’ career goals affects their engagement, satisfaction, and academic performance. We also examine how this personalization shapes learners’ perceptions of relevance and autonomy in their studies. By doing so, we contribute to the ongoing discussions about the role of AI in education and the importance of personalizing learning experiences for a future-oriented, skills-based economy.

This paper is organized as follows: Section 2 outlines the theoretical background, focusing on learning mechanisms and motivational theories relevant to personalized education. Section 3 reviews related work on adaptive learning systems and generative AI in educational contexts. Section 4 details the experimental setup, including study design, participants, and personalization methodology. In Section 5, we present the results of our study, including quantitative and qualitative findings. Finally, Section 6 discusses the implications of these results, addresses limitations, and outlines directions for future research. 



\section{Theoretical Background}

As a conceptual foundation, in this section, we provide an overview of key learning mechanisms and motivational theories relevant to personalized education, followed by a review of scenario-based methods in the context of career goal alignment.

\subsection{Fundamentals of Learning}

Modern learning science highlights that effective education requires more than content delivery--—it demands active engagement with material, long-term retention, and the ability to transfer knowledge to novel situations. Insights from neuroscience, psychology, and education theory offer a multidimensional understanding of how learning processes can be optimized.

From a neurobiological perspective, learning is based on the principle of \textit{neuroplasticity}, the brain’s ability to reorganize and strengthen neural pathways in response to new experiences. Mechanisms such as long-term potentiation in the hippocampus are critical for memory formation and consolidation \cite{kandel2014}. These processes are modulated by the dopaminergic system, which responds to novelty and reward, enhancing memory encoding when information is perceived as meaningful or emotionally salient \cite{lisman2005,shohamy2010}.

In parallel, \textit{multisensory learning}---the simultaneous engagement of visual, auditory, and kinesthetic channels---supports more effective knowledge integration. Studies show that multimodal input strengthens memory encoding and retrieval, especially when paired with emotional significance or real-world relevance \cite{shams2008}.

Psychological theories further support the design of effective learning environments. \textit{Behaviorism} emphasizes reinforcement and repetition, especially in early skill acquisition. \textit{Cognitivist} approaches prioritize active mental processing, with principles like the spacing effect and retrieval practice shown to improve long-term retention \cite{roediger2006}. The spacing effect refers to distributing learning over time rather than engaging in massed practice, which helps the brain consolidate information more effectively. Retrieval practice, on the other hand, involves actively recalling information from memory instead of simply rereading it, thereby strengthening memory traces and enhancing durable learning. Meanwhile, \textit{constructivist} theories argue that learners build knowledge through reflection, experience, and contextualized exploration \cite{vygotsky1978}.

Motivation is a key driver of sustained learning. The \textit{Self-Determination Theory} \cite{deci1985} identifies autonomy, competence, and relatedness as essential for intrinsic motivation. When learners feel in control, capable, and connected to meaningful goals, they are more likely to persist in challenging tasks. The \textit{Expectancy-Value Theory} \cite{eccles2002} complements this view, emphasizing that motivation increases when learners believe they can succeed and when they perceive the content as valuable. Finally, the \textit{Goal Orientation Theory} \cite{dweck2006} suggests that learners who adopt a growth mindset---viewing abilities as improvable---engage more deeply with complex material and embrace mistakes as learning opportunities.

These theories collectively indicate that learning is most effective when it is emotionally relevant, cognitively engaging, and motivationally aligned with the learner’s goals and identity.

\subsection{Personalized Learning Through Scenario-Based Methods}

To meet the evolving demands of the 21st-century workplace, learners must develop not only knowledge but also critical thinking, decision-making, and problem-solving skills. Scenario-based learning, which also includes problem-based learning, offers a powerful pedagogical model to improve these competencies. Instead of passive memorization, learners are immersed in realistic professional situations that require them to analyze, evaluate, and act—mirroring the complexity of real-world challenges \cite{pitorini2024}.

This type of learning promotes cognitive flexibility, as learners must consider multiple perspectives, adapt their strategies, and justify their decisions. It activates executive brain functions, particularly in the prefrontal cortex, supporting metacognition, planning, and reflective judgment \cite{miller2001}. Repeated engagement with such high-relevance scenarios also strengthens memory encoding through emotional salience and relevance to learners’ personal goals.

In higher education, when these scenarios are personalized to learners’ career goals, the pedagogical benefits are amplified. Learners are more likely to perceive the material as useful, increasing intrinsic motivation and willingness to engage. The integration of professional contexts fosters the transfer of learning--—the ability to apply knowledge in new, complex situations, which is central to employability and lifelong learning \cite{eccles2002}.

Scenario-based personalization also supports error-based learning. By encountering realistic mistakes in safe, simulated environments, learners can reflect on consequences and revise their thinking without real-world risks. This iterative process builds confidence, deepens understanding, and strengthens professional decision-making skills \cite{pitorini2024, mangelsWhyBeliefsIntelligence2006}.

Motivationally, aligning tasks with individual career aspirations satisfies the three core psychological needs defined by Self-Determination Theory: autonomy (choosing learning paths), competence (overcoming relevant challenges), and relatedness (seeing personal purpose in education) \cite{deci1985}. Moreover, when learners experience success in solving career-related problems, dopaminergic reward circuits are activated, reinforcing not only memory formation but also positive associations with learning \cite{schultz2016}.

In summary, career goal-oriented scenario-based learning offers a theoretically grounded, neurocognitively supported, and motivationally rich approach to education. It bridges the gap between abstract academic knowledge and professional application—a gap increasingly critical to address in AI-enhanced, future-ready learning environments.

\section{Related Work}

In recent years, research on AI in education has increasingly focused on adaptive systems that personalize learning to individual needs and goals. 

Previous work has shown that personalized learning environments—especially those simulating realistic decision-making—can foster critical thinking and knowledge transfer. For instance, \cite{pitorini2024} combined problem-based learning with AI-driven reflective questioning and reported significant improvements in learners’ analytical thinking. The study was conducted with ten tenth-grade high school students (ages 15–16) in Indonesia and focused on the biology topic of environmental change. Participants used an e-module structured around problem-based learning combined with Socratic dialogue, implemented over three 90-minute sessions. Their results emphasize the importance of tailoring educational content to relevant, context-specific scenarios. However, the generalizability of their findings is limited by a small sample size and short intervention period.

Other implementations of AI in higher education have focused on micro-adaptive learning platforms. \cite{baillifard2023} developed a GPT-based tutoring system that dynamically generated quiz questions and tracked learners' knowledge states. This led to a notable increase in exam performance, up to 15 percentile points higher compared to control groups. While promising, these approaches largely optimized content delivery and retrieval, rather than aligning with long-term career development goals.

Several projects, including DIMA~\cite{bucher2024} and SokratesT~\cite{hochschuleRheinWaal}, have explored how GenAI can support dialogic learning, i.e., building knowledge through open dialogue and collaborative interaction, and simulate workplace-relevant communication scenarios. These systems use AI to create authentic task environments (e.g., negotiations, critical incidents) and offer personalized feedback. Although learners reported increased engagement and perceived realism, most systems did not yet incorporate explicit learner goals such as job roles or career trajectories.

Large-scale empirical studies such as~\cite{moeller2024} have evaluated AI-based learning platforms at an institutional level. While their findings highlight measurable efficiency gains—e.g., a reduction in time-to-exam—the platforms primarily focused on logistical optimization, such as the improved organization and delivery of exam-relevant materials, rather than the content-level personalization examined in our study.

The use of GenAI for career goal-based personalization remains an emerging field. Recent reviews have called for more research into how adaptive systems can support identity-driven learning~\cite{kasneci2023, ng2024}. They stress the need for transparent, ethically grounded AI that not only adapts to learners' prior knowledge but also supports their long-term goals and professional identities. \cite{Holmes2021Ethics} highlight the need for community-wide ethical frameworks in AI in education, arguing that responsible innovation must go beyond improving learning outcomes to address fairness, accountability, and inclusivity in system design. Similarly, \cite{Baker2021Bias} document how biased training data and non-transparent algorithms can amplify existing disparities, particularly affecting marginalized groups in education.

In contrast to prior work, our study focuses specifically on the effects of aligning learning content with learners’ career goals using GenAI. We contribute to this growing field by empirically evaluating how such personalization affects learner engagement, satisfaction, and performance in a real-world university setting. Our results complement existing research and suggest that content aligned to individual career goals can substantially increase perceived relevance, deepen cognitive engagement, and support knowledge transfer to professional contexts.

\section{Experimental Setup}

To investigate the impact of career goal-based personalization in AI-driven learning environments, we conducted a controlled field study with learners from International University of Applied Sciences (IU). The study was not limited to a specific program or degree level and included bachelor's and master's students from a variety of disciplines. 
The primary goal was to evaluate whether aligning learning content with learners’ professional aspirations improves engagement, satisfaction, and learning efficiency.

\subsection{Study Design and Participants}

The study was implemented using Syntea, an AI-powered learning assistant developed by IU. Syntea is integrated into IU’s digital learning platform and provides personalized, conversational learning experiences based on generative language models. It combines a proprietary didactic framework with a personalization engine, enabling dynamic adaptation of content to individual learner profiles and goals. Core functionalities include Q\&A support, interactive exam preparation, and progress tracking, all designed to enhance student engagement and learning efficiency. For our study, Syntea was extended to align learning content with self-reported career aspirations, allowing for the delivery of profession-specific examples and scenarios.

We used a between-subjects A/B test design with two experimental conditions:

\begin{itemize}
  \item \textit{Career goal}: learners in this group received personalized learning scenarios tailored to their individual career goals.
  \item \textit{Traditional}: learners received standard, non-personalized content covering the same academic topics.
\end{itemize}

A between-subjects design where each participant experiences only one type of instructional approach was chosen to minimize potential biases and confounding variables that could arise from carryover effects. 

A total of 4,003 learners participated in the study over six weeks during the winter semester of 2024. The learners were randomly assigned to one of the two groups upon accessing the AI-based learning platform to ensure that participant allocation did not influence the study results. Table \ref{tab:participants} compares the participants in our study, detailing the number of learners and sessions across both groups, along with average sessions per learner and their standard deviations.

\begin{table}[h]
\centering
\label{tab:participants}
\caption{Participants of our Study.}
\begin{tabular}{@{}lrrr@{}}
\toprule
\textbf{Condition} & \textbf{\# Learners} & \textbf{\# Sessions} & \textbf{\# Sessions per Learner (SD)}  \\
\midrule
Career Goal & 1,700 & 8,360 & 4.92 (8.15) \\
Traditional     & 2,303 & 10,668 & 4.63 (7.32) \\
\midrule
\textbf{Total} & 4,003 & 19,028 & 4.76 (7.69) \\
\bottomrule
\end{tabular}

\end{table}

The \textit{traditional} condition had more participants and total sessions than the \textit{career goal} condition. This might be due to higher entry complexity and the initial requirement to provide career-related input. On average, participants in the \textit{career goal} condition attended slightly more sessions than those in the \textit{traditional} condition.

The AI system used in this experiment was a generative language model integrated into the university's existing learning platform, Syntea. The platform includes a proprietary didactic framework and personalization engine developed specifically to adapt content to individual learner profiles and goals. In our study, the system provided learners with conversational, scenario-based learning units derived from their course materials. In the \textit{career goal} condition, the system dynamically adapted the prompts and examples to reflect learners’ previously submitted career goals (e.g., ``project manager in IT'' or ``emergency nurse'').

Personalization was achieved through a combination of metadata (such as intended job role) and internal prompt adaptation methods, proprietary to Syntea. Each session consisted of open-ended questions, contextually embedded problem scenarios, and reflective follow-up prompts. Upon completion, learners received performance feedback and suggestions for further improvement. The described functionalities are core components of the Syntea platform, which was further extended with the career goal functionality that we analyze in this study.

\subsection{Data Sources}
To gain a comprehensive understanding of learner engagement and performance, data for the study were drawn from two main sources:
\begin{enumerate}
  \item {Learning Interaction Logs}: Automatically recorded data from the platform, including session length, number of interactions, and completion rates.
  \item {Academic Performance Data}: Exam results and study durations (time between course enrollment and exam completion) were obtained from the university’s data warehouse.
\end{enumerate}

Additionally, learner satisfaction was assessed using the Net Promoter Score~(NPS) and collected via an in-platform survey after each session. The NPS is a standardized metric that measures the likelihood of learners recommending the product to others, serving as an indicator of overall satisfaction and perceived value. Participants were asked the question ``How likely are you to recommend our session to a friend on a scale of 0 to 10?''. 

\subsection{Key Metrics for Analysis}

To evaluate the effects of personalization, we conducted independent two-sample \textit{t}-tests comparing both groups across key metrics:

\begin{itemize}
  \item \textit{Learner Engagement}: Measured by mean and median session duration, number of sessions per learner, and the percentage of successful sessions.
  \item \textit{Learner Satisfaction}: Based on average NPS scores.
  \item \textit{Study Efficiency}: Assessed by study duration and exam scores.
\end{itemize}

We also performed a qualitative analysis of open-ended \textit{learner feedback} to identify perceived strengths and weaknesses of the career goal personalization approach.

All data was anonymized before analysis. Participation in the learning tool was voluntary and part of a regularly available study support system. The study complied with institutional data protection policies and relevant ethical standards. Participants were informed about the Syntea Data Privacy Policy and actively consented to it before using the learning tool.

\section{Results}

This section presents the results of the experimental study, focusing on the effects of career goal-based personalization on learner engagement, academic performance, and perceived learning value. Both quantitative and qualitative data were analyzed to assess the impact of aligning AI-generated content with learners’ professional aspirations.

\subsection{Data Selection}

To ensure sufficient exposure, only participants who engaged in at least five learning sessions were included in the quantitative analysis. 
This threshold was based on the course structure of IU International University, where each course consists of six units, each covering five topics. Five sessions represent the completion of one full unit and therefore indicate a meaningful level of progress.
This criterion helped to filter out casual learners and ensured that only those with meaningful engagement were analyzed. Table \ref{tab:participantsFinalAnalysis} shows the number of participants included in our quantitative analysis. 

\begin{table}[h]
\centering
\caption{Participants Included in Quantitative Analysis (Five or More Sessions).}
\label{tab:participantsFinalAnalysis}
\begin{tabular}{@{}lrr@{}}
\toprule
\textbf{Condition} & \textbf{\# Learners} & \textbf{\# Sessions}  \\
\midrule
Career Goal & 445 & 6,312\\
Traditional     & 607 & 7,920  \\
\midrule
\textbf{Total} & 1,052  & 14,232 \\
\bottomrule
\end{tabular}
\end{table}

About a quarter of the learners in each group participated in five or more sessions. Specifically, 26.2\% of the \textit{career goal} group (445 out of 1,700 learners) engaged in at least five sessions. Similarly, 26.4\% of learners in the \textit{traditional}  group (607 out of 2,303 learners) attended five or more sessions. This suggests a comparable level of commitment in both groups.

Focusing only on the participants who attended five or more sessions, a total of 1,052 learners collectively completed 14,232 sessions. The groups are split as follows: 445 learners, representing 42.3\% of all learners, are in the \textit{career goal} group, while 607 learners (57.7\%) are in the \textit{traditional} group. With 6,312 sessions, learners in the \textit{career goal} group contributed approximately 44.4\% of all sessions (14,323 sessions in total), while participants in the \textit{traditional} group completed 7,920 sessions, accounting for 55.6\%. Although the \textit{career goal} group has fewer learners, the proportion of total sessions they contribute is similar to their proportion of the total number of learners.

\subsection{Data Analysis}
In the following subsections, we present the results of the analysis with regard to the above-defined key metrics.

\subsubsection{Learner Engagement}

To measure the learner engagement, we used different metrics, which are displayed in Table \ref{tab:kpm}.

\begin{table}[h]
\centering
\caption{Comparison of Key Performance Metrics.}
\label{tab:kpm}
\begin{tabular}{@{}lrrr@{}}
\toprule
\textbf{Metric} & \textbf{Traditional} & \textbf{Career Goal} & \textbf{\% Diff.}  \\
\midrule

Mean session duration (sec) & 1,385.75 & 1,457.89 & +5.21\%  \\
Median session duration (sec) & 813.76 & 916.46 & +12.62\%  \\
Mean \# sessions per learner & 13.05 & 14.18 & +8.66\%  \\
Successful sessions (\%) & 82.41 & 80.81 & -1.94\%  \\
\bottomrule
\end{tabular}
\end{table}

On average, learners in the \textit{career goal} group ($M_c=1,457.89$, $SD_c=1,186.46$) spent more time per session than those in the control group ($M_t=1,385.75$, $SD_t=1,352.12$), reflecting a 5.12\% increase. However, the difference is statistically not significant ($t(df)=-0.900$, $p = .3684$). 
Both groups show substantial variation in the time spent per session, as indicated by their standard deviations. The \textit{traditional}  group, with a higher standard deviation, demonstrates even greater variability in how much time different learners spent per session.

Learners in the \textit{career goal} group attended an average of 14.18 sessions ($SD_c=11.62$), while learners in the \textit{traditional} group attended an average of 13.05 sessions ($SD_t=10.24$). 
Although the average number of sessions per learner was slightly higher in the \textit{career goal} group (+8.66\%), this difference was not statistically significant ($t(1050)=-1.679$, $p = .0934$). 
The variability in session attendance was somewhat greater in the Career Goal group, as indicated by the higher standard deviation (11.62 vs. 10.24). However, the overall difference in variability between the two groups is not particularly large. 

Finally, 82.41\% of sessions were successful in the \textit{traditional} group, compared to 80.81\% in the \textit{career goal} group. This represents a 1.94\% lower success rate in the \textit{career goal} group. However, with a $p$-value of 0.1525, this difference is not statistically significant ($t(1050)=1.432$). A session was considered successful if the underlying model detected that the learner had worked through all relevant concepts within the topic. In such cases, the learning platform generated a session summary and prompted the learner for feedback, indicating successful completion of the learning unit.


\subsubsection{Learner Satisfaction}

\begin{figure}[htbp]
    \centering
    \includegraphics[width=0.5\linewidth]{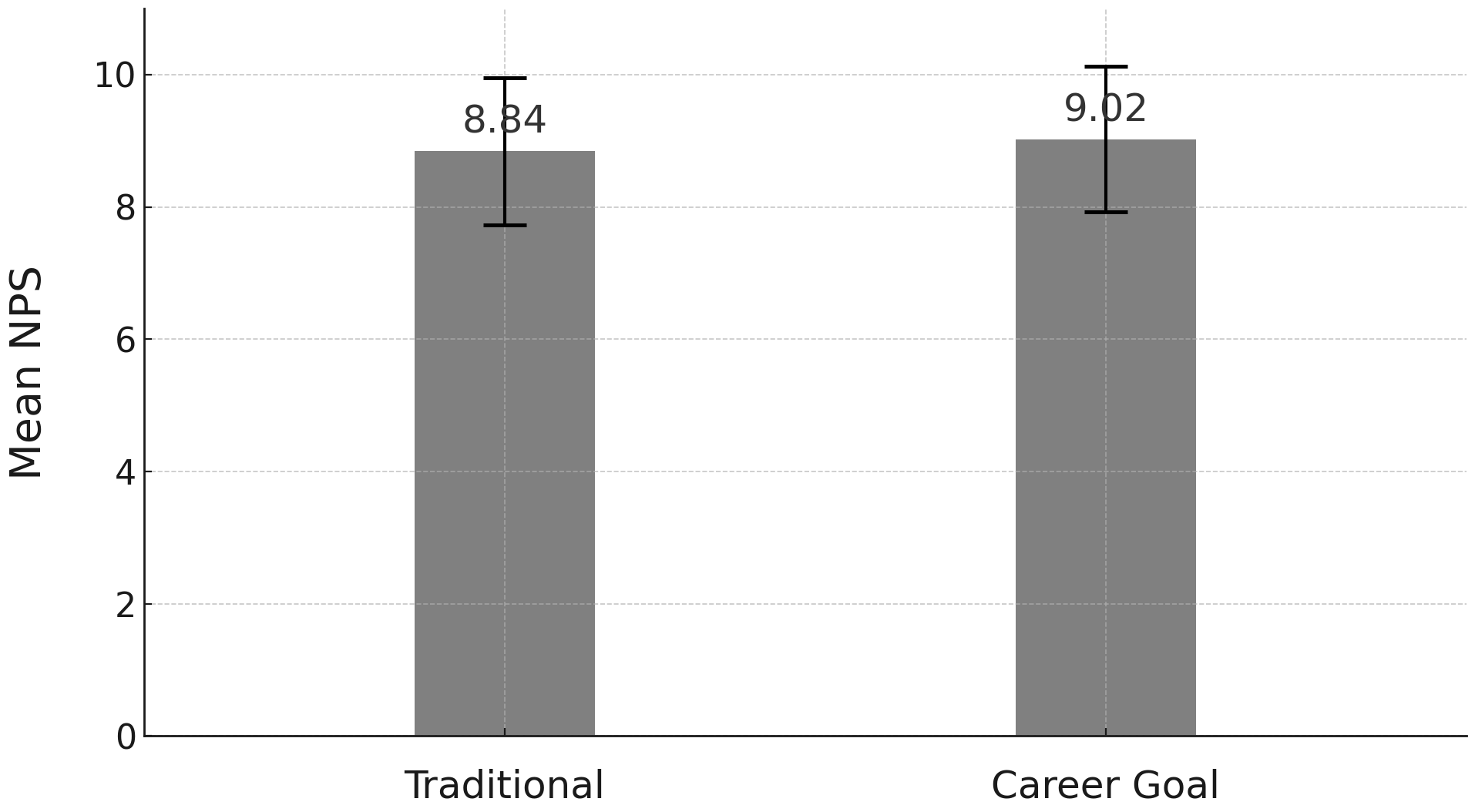}
    \caption{Mean NPS for \textit{Traditional} and \textit{Career Goal} Group (Error Bars Indicate Standard Deviation).}
    \label{fig:NPS}
\end{figure}

To analyze the learner satisfaction, we analyzed the NPS for both groups. The results are depicted in Figure \ref{fig:NPS}.

While the NPS ranges from 0 to 10, a score of around 9 is considered high and indicates so-called promoters. These individuals are highly likely to recommend the learning approach to others, suggesting a strong level of satisfaction and endorsement among participants with the dialogic learning based on GenAI.

The average NPS per learner was slightly higher in the \textit{career goal} group ($M_c = 9.02$, $SD_c = 1.10$) compared to the \textit{traditional} group ($M_t = 8.84$, $SD_t = 1.11$). This difference was statistically significant, $t(942) = -2.448$, $p = 0.0145$, with a small effect size ($d = -0.161$), indicating that learners in the \textit{career goal} version reported slightly greater satisfaction. The similar standard deviations in both groups suggest that the variation in satisfaction levels was consistent across both groups, reinforcing the reliability of this finding.

\subsubsection{Study Efficiency}

To analyze the relationship between personalized learning and academic outcomes, we used the exam scores and study duration. The results are illustrated in Table \ref{tab:exam_results}. It is important to note that not all learners completed their courses within the analyzed study period. As a result, the analysis of study efficiency is based on a subsample of 241 learners in the \textit{career goal} group and 584 learners in the \textit{traditional} group.

\begin{table}[h]
\centering
\caption{Comparison of Exam Performance and Study Duration.}
\label{tab:exam_results}
\begin{tabular}{@{}lrrr@{}}
\toprule
\textbf{Metric} & \textbf{Traditional} & \textbf{Career Goal} & \textbf{\% Diff.} \\
\midrule
Exam results (\%) & 77.73 & 78.46 & $+0.94$\% \\
Study duration (days) & 160.74 & 148.66 & $-7.51$\% \\
\bottomrule
\end{tabular}
\end{table}

Exam scores were nearly identical for both groups. Even though these findings do not indicate significant gains in academic performance, the reduced study duration in the personalized group may point to increased learning efficiency. Specifically, learners in the \textit{career goal} group completed their courses more quickly from enrollment to exam compared to the control group (7.51\% faster). However, this difference did not reach statistical significance ($t(514.88) = -.851$, $p = .3955$).

\subsubsection{Learner Feedback}

In addition to the quantitative analysis, we also performed a qualitative analysis of the open-ended responses from the \textit{career goal} group. For this analysis, we analyzed the feedback of all participants, including those who participated in fewer than five sessions. The analysis followed a two-step inductive and fully manual approach. In the first step, all responses were reviewed to identify overarching response types. Based on this open coding, four main categories emerged: positive feedback, suggestions for improvement, negative feedback, and technical issues. Each comment could be assigned to one or more categories to reflect its complexity. In the second step, we manually examined the content within each category to identify recurring themes and frequently mentioned topics. We collected a total of 1,660 responses from 500 different learners, which we grouped based on their content. Overall, 74\% of comments were positive in tone, 22\% suggested areas for improvement, and only 5\% were explicitly negative. Additionally, 4\% of the responses comprised technical issues. Responses addressing multiple themes were coded into each corresponding category to ensure a comprehensive analysis. Consequently, category percentages exceed 100\%, as responses were counted in more than one category.

\begin{figure}[htbp]
    \centering
    \includegraphics[width=1\linewidth]{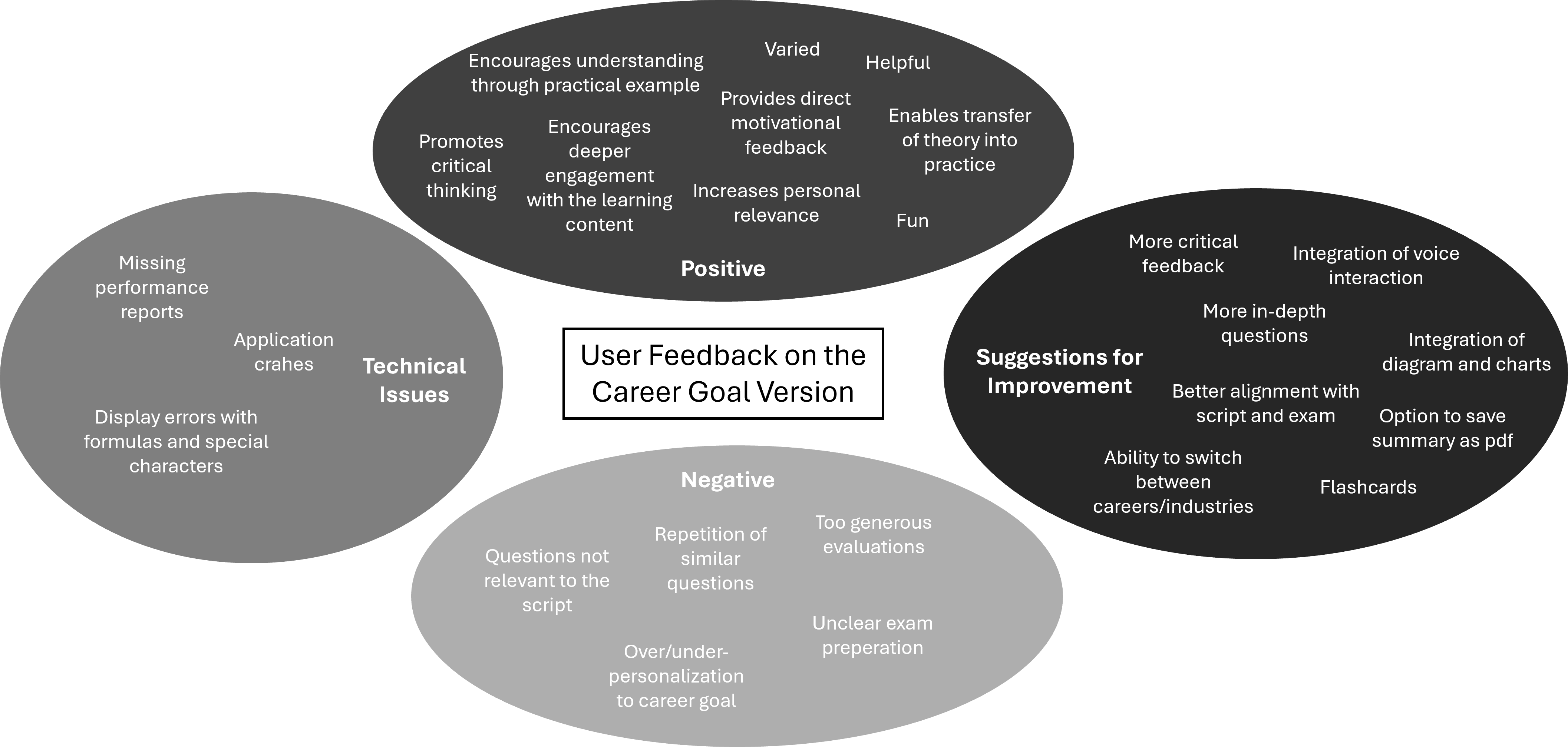}
    \caption{Summary of Learner Feedback.}
    \label{fig:feedback}
\end{figure}

Figure \ref{fig:feedback} summarizes the learner feedback for the \textit{career goal} group.

Many learners emphasized that the personalized content felt directly applicable to their intended careers, which increased their motivation and attention. Comments such as ``This is the first time I saw myself in the material'' and ``Finally, something that connects to my future job'' highlight a strong sense of identification. Learners described the content as ``more real'', ``less abstract'', and ``professionally meaningful''. Several noted that the relevance made them ``want to continue learning even after the session ended''.
Learners frequently reported that the career-aligned scenarios prompted deeper thinking. Rather than simply recalling information, learners engaged in role-based problem-solving and critical decision-making. One learner remarked: ``I had to really think about what I would do in this situation at work — not just remember theory.'' Others highlighted that reflecting on realistic challenges in their own field helped them retain knowledge more effectively.

The scenario-based dialogue format was widely appreciated. Learners described the sessions as ``dialogue-driven'', ``alive'', and ``more like a simulation than a lesson''. This format was seen as a break from typical static learning tools. Several learners specifically mentioned that the format helped them articulate thoughts more clearly and improved their confidence in workplace-relevant reasoning.

While most feedback was positive, 22\% of learners expressed constructive criticism. Some wished for greater alignment between the personalized content and exam preparation, suggesting that ``some questions were too far from what is tested''. Others requested more depth in domain-specific content or greater variability in scenario types. A few mentioned inconsistencies in the relevance of the example scenarios to their chosen careers, indicating potential areas for refinement in the personalization algorithm.

A minority of comments (4\%) referenced technical issues such as loading delays or formatting errors. These were mostly isolated and not indicative of systemic problems. Nonetheless, they underscore the importance of robust system performance when deploying AI-driven tools in real educational contexts.

In summary, learner feedback strongly supports the use of career goal-based personalization as a way to increase emotional connection, relevance, and engagement in digital learning environments. The comments suggest that personalized scenarios do more than improve satisfaction — they promote reflective, contextualized thinking aligned with learners’ professional identities.

\subsection{Discussion}

In our study, we explored the impact of career goal-based personalization in an AI-driven learning environment, focusing on learner engagement, satisfaction, and academic performance. Our findings show that aligning educational content with learners’ professional aspirations substantially increases engagement metrics and perceived learning value. Participants in the personalized condition reported higher satisfaction, spent more time in sessions, and expressed a stronger connection to the material through qualitative feedback. Learners in the \textit{career goal} group highlighted potential areas for improvement, offering valuable insights for refining the learning approach. 

Although exam performance did not differ significantly between groups, the reduced average study duration suggests that personalization may support more efficient learning trajectories. These findings reinforce theoretical models such as\textit{ Self-Determination Theory} and \textit{Expectancy-Value Theory}, which posit that autonomy and goal relevance are central to sustained motivation and effective learning.

By integrating generative AI with goal-based scenario design, this study contributes to the growing field of adaptive educational systems and highlights the potential of identity-aligned personalization as a meaningful innovation in higher education.

Several limitations should be noted. First, the experimental design relied on self-reported career goals provided at the start of the course, which may not fully reflect learners’ evolving professional identities or short-term learning goals. Moreover, the depth and specificity of the reported goals varied substantially across participants, limiting the precision of the personalization logic. This self-report-based approach also introduced a potential self-selection bias, as learners who chose to work with Syntea may have had a stronger intrinsic motivation, more digital affinity, or clearer professional objectives compared to those who did not.
Second, while the sample size was large, the study was conducted within a single institutional and cultural context, limiting generalizability to other educational settings or disciplines. As the sample consisted exclusively of distance learning students, who are typically more accustomed to digital tools and autonomous learning, the findings may not fully translate to learners in on-campus learning environments. Moreover, potential differences between individual study programs were not examined. Future research should address these aspects to better understand how subject-specific contexts influence the effectiveness of personalized learning approaches.

Third, the personalization logic—although effective—was based on relatively simple prompt conditioning and did not yet incorporate deeper models of learner profiles, prior knowledge, or emotional state. Furthermore, the assessment of academic success focused solely on exam outcomes and study duration, which may not capture deeper learning or skill acquisition.

Finally, the study treated exam outcomes and study duration as independent indicators of academic success. However, these dimensions may be interrelated---for instance, learners aiming for higher grades might deliberately invest more time, while those prioritizing speed may accept lower performance. This potential trade-off was not explicitly accounted for in the current analysis.

\section{Conclusion and Future Work}

In this section, we summarize the main findings of our study, reflect on their implications for AI-supported learning design, and outline directions for future research.

Overall, the results demonstrate that career goal-based personalization using GenAI has a measurable, though nuanced, impact on the learning experience. 
The combination of quantitative and qualitative analyses proved particularly valuable in uncovering its full impact. While the quantitative metrics revealed positive trends in learner engagement, satisfaction, and study efficiency, the qualitative feedback offered deeper insights into learners’ personal experiences and the emotional relevance of the content. 

This study makes an important contribution to the research on AI in education technologies. Our findings demonstrate that GenAI, when applied purposefully, can do more than streamline learning processes--—it can enrich them. Learners not only responded positively to the personalized approach but also engaged deeply with the material, suggesting that GenAI has the potential to support not just efficiency, but meaningful, learner-centered education.




However, several areas require further investigation. Personalization systems based on GenAI carry inherent risks of algorithmic bias and overfitting, particularly when relying on limited or static user input to generate context-specific content. Such risks must be addressed to ensure alignment with diverse learner needs and to avoid reinforcing narrow assumptions about professional identities.

Moreover, learners with undecided or evolving career goals may not benefit equally from goal-specific personalization. For these individuals, rigid alignment with predefined pathways may hinder exploration and personal development. Future personalization approaches should therefore incorporate flexible, exploratory learning pathways that support identity formation and dynamic goal adjustment.

Future research should also explore the longitudinal effects of career goal-based personalization, particularly in terms of long-term knowledge retention, job preparedness, and learner autonomy. More sophisticated learner modeling—incorporating real-time feedback, motivation tracking, and dynamic goal adjustment—--could further enhance personalization and learning outcomes.

Moreover, cross-institutional studies in diverse learning contexts are needed to validate the findings and assess scalability. Investigating how personalization interacts with other pedagogical strategies, such as peer collaboration or competency-based assessment, would also offer valuable insights.

Finally, ethical considerations around goal profiling, algorithmic transparency, and student agency must be addressed to ensure that personalized AI systems are not only effective but also equitable and trustworthy.

\section{Ethical Impact Statement}
This study was conducted in accordance with ethical standards and data protection regulations. All learner data collected during the research was fully anonymized prior to analysis, ensuring that no personally identifiable information could be traced back to individual participants at any point in the research process. Participation in the AI-powered learning tool was entirely voluntary, with students free to opt in or out without any academic consequences. 
Syntea is part of an ongoing academic support program, and its use requires informed consent as defined by its internal data privacy policy. 
No sensitive or high-risk personal data was collected. 
The system was designed with privacy-by-design principles to ensure compliance with data governance best practices.
Overall, the research protocol minimized risk to participants while maximizing educational benefit and scientific rigor.


\bibliography{20-bibliography}


\end{document}